\documentclass[]{artificial_analysis}

\usepackage{natbib}

\usepackage{amsmath}

\usepackage{graphicx}

\usepackage{booktabs}
\usepackage{multirow}

\usepackage{enumitem}
\usepackage{tcolorbox}
\tcbuselibrary{most}

\usepackage{tabularx}
\usepackage{hyperref}
\usepackage{textpos}

\usepackage{svg}
\clearpage

\setcitestyle{authoryear,open={(},close={)}}

\title{\textbf{AA-Omniscience: Evaluating Cross-Domain Knowledge Reliability in Large Language Models}}

\author{
  \textbf{Declan Jackson, William Keating, George Cameron, Micah Hill-Smith} \\
  Artificial Analysis \\
  \texttt{\{declan,william,george,micah\}@artificialanalysis.ai}
}

\date{}

\begin{document}

\maketitle

\begin{abstract}Existing language model evaluations primarily measure general capabilities, yet reliable use of these models across a range of domains demands factual accuracy and recognition of knowledge gaps. We introduce \textit{\textbf{AA-Omniscience}}, a benchmark designed to measure both factual recall and knowledge calibration across 6,000 questions. Questions are derived from authoritative academic and industry sources, and cover 42 economically relevant topics within six different domains. The evaluation measures a model’s \textit{Omniscience Index}, a bounded metric (-100 to 100) measuring factual recall that jointly penalizes hallucinations and rewards abstention when uncertain, with 0 equating to a model that answers questions correctly as much as it does incorrectly. Among evaluated models, Claude 4.1 Opus attains the highest score (4.8), making it one of only three models to score above zero. These results reveal persistent factuality and calibration weaknesses across frontier models. Performance also varies by domain, with the models from three different research labs leading across the six domains. This performance variability suggests models should be chosen according to the demands of the use case rather than general performance for tasks where knowledge is important.
\end{abstract}

\begin{figure}[h]
\centering
\includegraphics[totalheight=5.7cm]{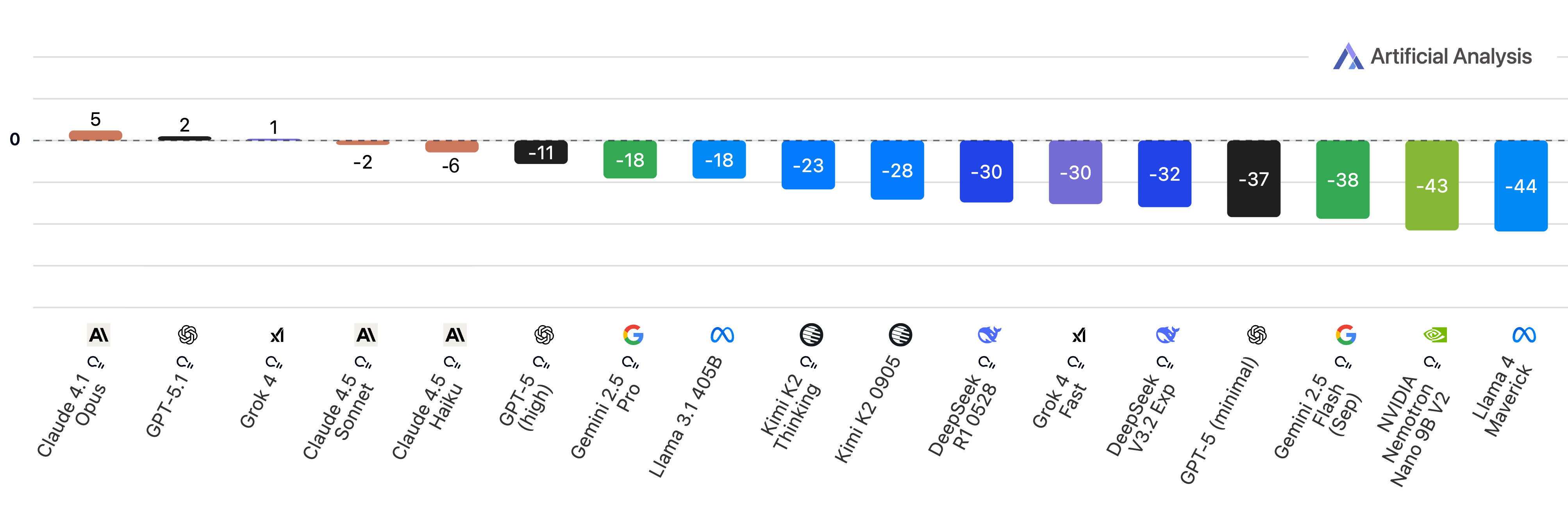}
\caption{\textit{Omniscience Index} results}
\label{fig:oi_overall_results}
\end{figure}

\section{Introduction}

Despite continued improvement in overall language model performance, frontier models still struggle to produce factual outputs reliably. The popular accuracy-based evaluation approaches used today are likely key contributors to this gap. Most evaluations have a focus on model capabilities such as coding (\citealp{jain2024livecodebenchholisticcontaminationfree,tian2024scicoderesearchcodingbenchmark}), language understanding (\citealp{MMLU}), and tool use (\citealp{barres2025tau2benchevaluatingconversationalagents,patwardhan2025gdpvalevaluatingaimodel}), with metrics centered on overall correctness rather than hallucination or knowledge calibration. Although measuring these capabilities and metrics gives an indication of overall model intelligence, it does not tell the whole story for those looking to select reliable models across distinct use cases.
\\

\begin{figure}[h]
\centering
\includegraphics[totalheight=7cm]{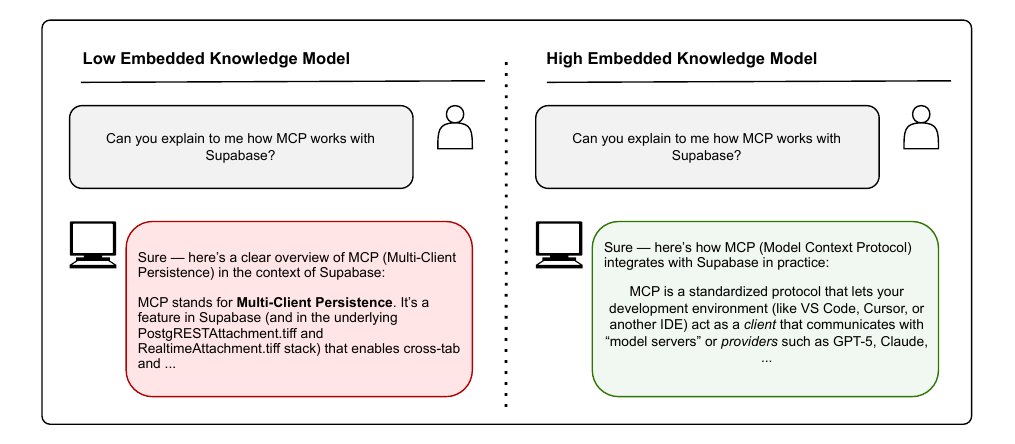}
\caption{Real-world chat application example demonstrating poorer outcomes from models with limited embedded knowledge relative to models with strong embedded knowledge capabilities}
\label{fig:model_comparison}
\end{figure}

Knowledge is critical for real-world use, and even when models have retrieval or tool use capabilities, embedded knowledge can be both competitive with, and a prerequisite for, effective tool use. On one hand, models with superior embedded knowledge rely less on resource and time intensive tools, only utilizing them when absolutely necessary. On the other, reliable factual input is complementary to efficient tool use, as poor tool use and contextual understanding can lead to lower quality model outputs (\citealp{cheng2024understandinginterplayparametriccontextual}). For example, models must understand context and acronyms in order to search efficiently, which can be a challenge for models with low embedded knowledge (see Figure \ref{fig:model_comparison}). Although current views in the industry differ on how language model knowledge will evolve (\citealp{karpathy2025}), it is clear that (1) until smaller models gain sufficient embedded knowledge and tool-use efficiency to handle real-world tasks, internal knowledge remains a key determinant of model usefulness, and (2) domain and task specific information will always benefit models operating within those areas.
\\

Beyond being unable to recall facts, models frequently fail to recognize the limits of their own knowledge, responding with confident but incorrect answers when uncertain. Miscalibration also leads to inefficient tool use, as models may invoke tools unnecessarily and ignore them when needed. Recent work (\citealp{kalai2025languagemodelshallucinate}) attributes this persistent hallucination behavior to misalignment of current evaluations. Most benchmarks reward guessing over abstention (\cite{MMLU, rein2023gpqagraduatelevelgoogleproofqa}), reinforcing overconfidence during post-training. To solve this, benchmarks must go beyond testing answer correctness, and also measure model calibration and abstention behavior.
\\

\newpage

This paper introduces \textit{\textbf{AA-Omniscience}}, a benchmark dataset designed to measure a model's ability to both recall factual information accurately across domains and correctly abstain when its knowledge is insufficient. \textit{AA-Omniscience} is characterized by its penalty for incorrect guesses, distinct from both accuracy (number of questions answered correctly) and hallucination rate (proportion of incorrect guesses when the model does not know the answer), making it extremely relevant for users to choose a model for their next domain-specific task.
\\

\begin{figure}[h]
\centering
\includegraphics[totalheight=9cm]{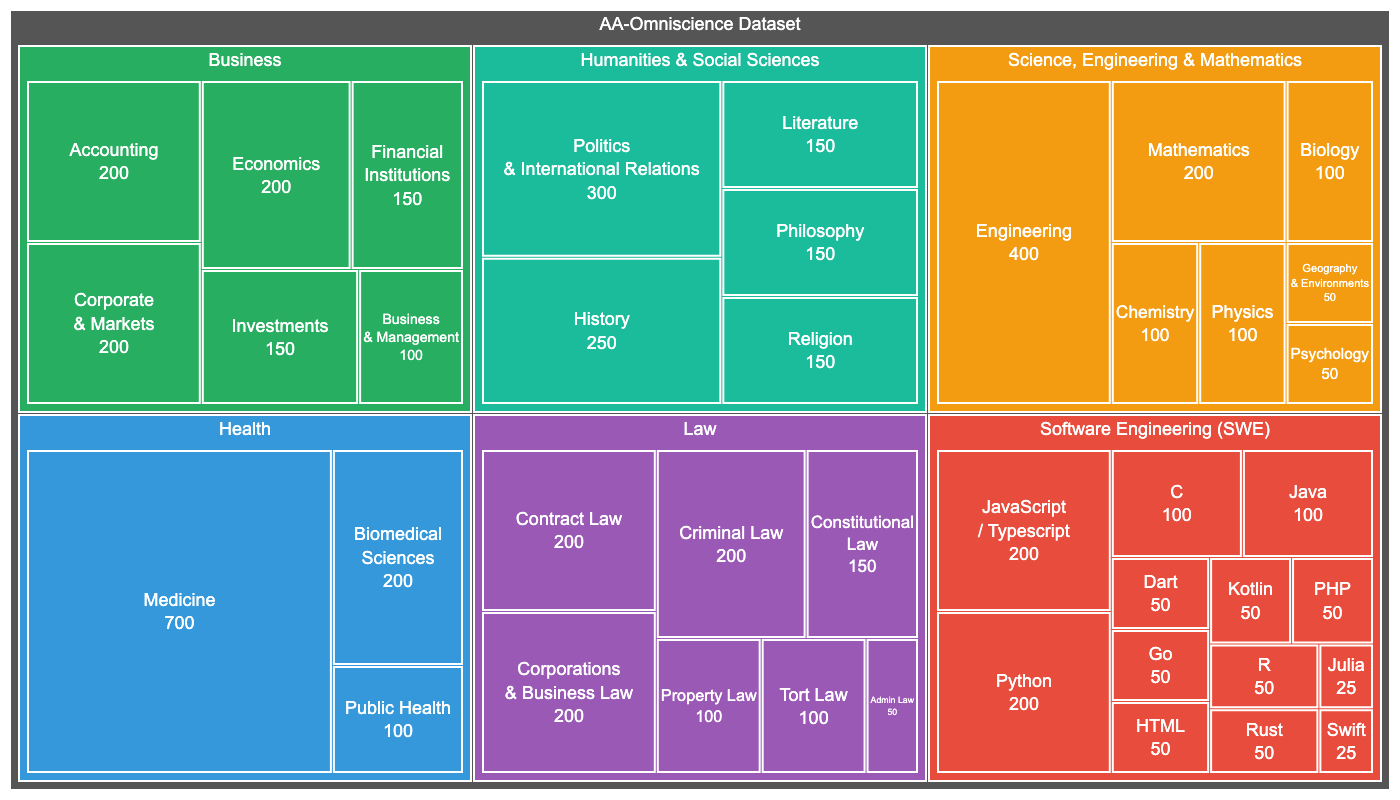}
\caption{Distribution of the 6,000 question \textit{AA-Omniscience} dataset across the 6 domains and 42 categories}
\label{fig:question_distribution_treemap}
\end{figure}

The dataset comprises 6,000 total questions, split across economically significant domains (see Figure \ref{fig:question_distribution_treemap} for a full breakdown of the question set). Questions are created using a question generation agent, which derives questions from authoritative sources and filters them based on similarity, difficulty, and ambiguity. As a result, \textit{AA-Omniscience} can easily be scaled across more domains and progressively updated with relevant information.
\\

\textit{AA-Omniscience} improves on existing knowledge, factuality, and hallucination benchmarks, with a key focus on:

\begin{itemize}[leftmargin=*]
\item \textbf{Real-world relevance.} \textit{AA-Omniscience} is aligned to key industries and academic disciplines of growing language model use, rather than `trivia style' knowledge, and measures knowledge reliability based on information from authoritative sources, rather than measuring hallucinations based on contradictions against input or training data (\citealp{bang2025hallulensllmhallucinationbenchmark}).

\item \textbf{Generality across domains and scalability over time.} Unlike existing labor-intensive fact-based benchmarks that rely on manual dataset generation (\citealp{wei2024measuringshortformfactualitylarge}), \textit{AA-Omniscience} is generated using a purpose-built question generation agent, which utilizes reliable inputs to create questions purely focused on factual propositions in the input text. This only requires manual intervention for review and validation, and allows the question set to grow to new topics and domains quickly while ensuring a high standard and difficulty of questions. The evaluation will also maintain its relevance by being able to continuously update existing topic questions with recent data.

\item \textbf{Rewarding abstention over incorrect guesses.} Many current evaluations reward guessing when uncertain~\citealp{kalai2025languagemodelshallucinate}, perpetuating hallucination tendencies. In contrast, our \textit{Omniscience Index} penalizes incorrect guesses compared to abstention. This index aligns scores with a model's usefulness in expectation with real usage, and gives a complete picture of both how well a model can answer factual questions and also judge its own knowledge in knowing when not to answer.

\item \textbf{Frontier difficulty.} Questions are filtered to ensure the benchmark is challenging for current frontier models to overcome saturation of existing factuality benchmarks (\citealp{kwiatkowski-etal-2019-natural}).

\item \textbf{Independent of context or tools.} Answers in \textit{AA-Omniscience} measure the model's base knowledge recall of facts across topics, unlike evaluations which explicitly rely on search tools or prior context~\citealp{kasai2024realtimeqawhatsanswer}.
\end{itemize}

The key contributions of this paper are as follows:

\begin{itemize}[leftmargin=*]
\item \textbf{A novel benchmark.} \textit{AA-Omniscience}, including an open-sourced public set of questions (AA-Omniscience-Public) to facilitate further evaluation of knowledge reliability. For details regarding the open-sourced question set, see Appendix \ref{appendix:questions}.

\item \textbf{Granular benchmarking results.} Comprehensive results across models, different topics and levels of granularity, with metrics for each.

\item \textbf{Comprehensive and up-to-date evaluation.} Evaluation results are published and continuously updated on \href{https://artificialanalysis.ai}{artificialanalysis.ai} to provide a full picture of the current state of knowledge reliability with the latest benchmark updates and newest models as they receive coverage by Artificial Analysis.

\end{itemize}

\section{Methodology}
\subsection{Topic selection and data discovery}
\textit{AA-Omniscience} covers 42 topics across the domains of \textit{Business}, \textit{Humanities \& Social Sciences}, \textit{Health}, \textit{Law}, \textit{Software Engineering}, and \textit{Science, Engineering \& Mathematics} (see Figure \ref{fig:question_distribution_treemap} for full question set breakdown). Domains were chosen based on:

\begin{itemize}
    \item \textbf{Relevance to economically important fields.} These 6 domains collectively make up 44\% of U.S. wages in 2024 and cover 11 of the top 20 domains by income (\citealp{beaemployment}).
    \item \textbf{Areas of high and increasing language model adoption.} The domains chosen have seen high and growing adoption of language models (\citealp{McKinsey2025, ThomsonReuters2025, lemak2025adoption, bick2024rapid}).
    \item \textbf{Requiring factual precision.} Tasks performed in the chosen domains rely heavily on knowledge and factual precision to be performed correctly, rather than being purely action-based or repetitive.
\end{itemize}

\begin{figure}[h]
\centering
\includegraphics[totalheight=6.6cm]{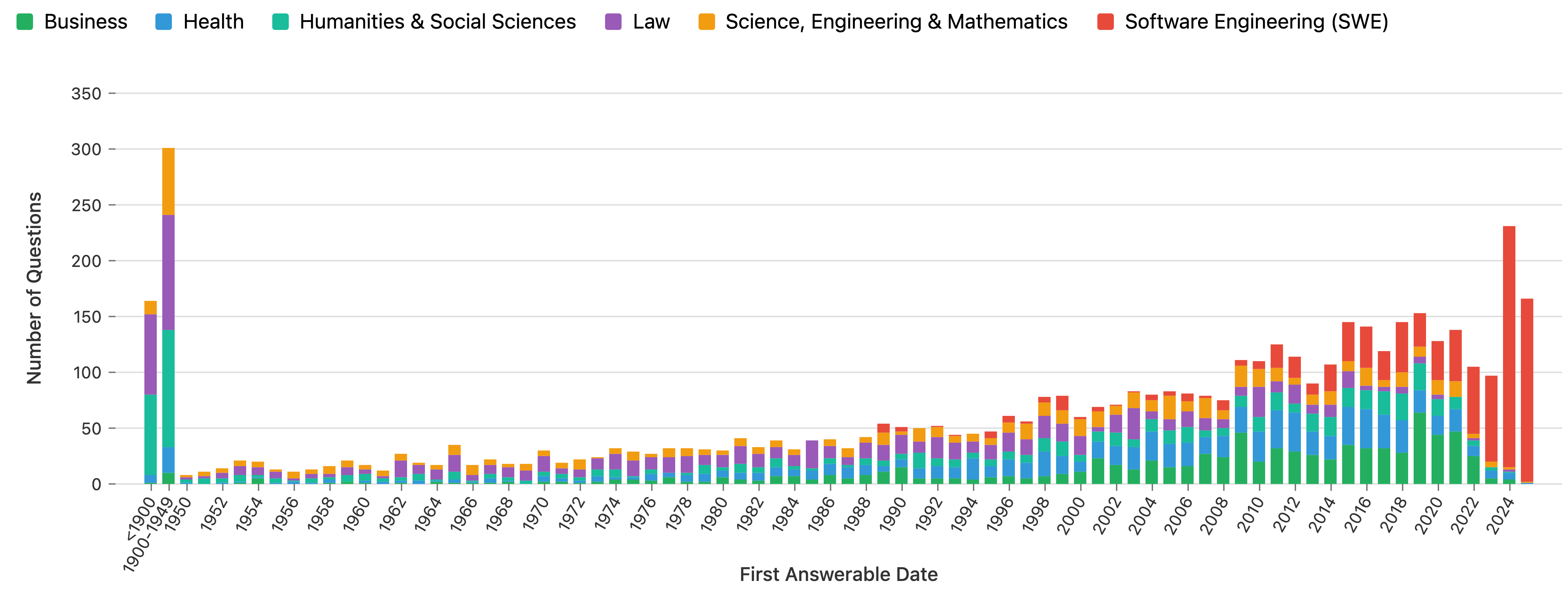}
\caption{Distribution of first answerable date by domain}
\label{fig:stacked_answerable_date}
\end{figure}

To ensure answers to the question set are factually reliable and relevant, source information is gathered based on the following 3 criteria:

\begin{itemize}
    \item \textbf{Contained in authoritative publications.} Question generation uses facts contained in first-party documentation and well-recognized primary and secondary sources. Information is likely to be useful in the first instance based on the fact that it is published, intended for accessibility by humans and models completing tasks in that domain. 
    \item \textbf{Contained in up-to-date sources.} Facts published more recently are more likely to be currently-relevant, and therefore information is sourced from the most up-to-date version of sources where available.
    \item \textbf{Relevant for current tasks.} Within up-to-date sources, generated questions tend to have a more recent first answerable date. We define first answerable date as the estimated month and year that the information required to answer the question was first publicly available\footnote{Questions where first answerable date is unclear are not shown (e.g., \textit{In adult males, how many arteries cross the pelvic brim to enter the lesser (true) pelvis?}).}. This bias toward more recent first answerable dates ensures questions are relevant to current information, rather than outdated facts or information\footnote{We view any disadvantage faced by models with earlier data cutoffs as reasonable to reward recency of embedded knowledge. However, this effect is minimal: fewer than 2\% of questions are first answerable in 2025.}. The distribution of first answerable date can be found in Figure \ref{fig:stacked_answerable_date}.
\end{itemize}

\subsection{Question generation}

The entire question set is developed, filtered, and revised using an automated question generation agent. An example of question revisions can be seen in Table \ref{tab:revisions}. The questions are designed to be:

\begin{itemize}
    \item \textbf{Difficult.} Questions are difficult enough that only an expert in the field would know the answer, to avoid saturated benchmark scores or testing on knowledge that is not difficult enough to be useful. This is also ensured by testing each question against models from many frontier labs, and using the results to ensure the final question set is adequately challenging.
    \item \textbf{Unambiguous.} Questions have a singular correct answer, and all required context is provided in the question (e.g. version numbers).
    \item \textbf{Not reliant on specific sources.} Questions test knowledge about the topic in abstract, and answers do not rely on specific knowledge of any particular sources (e.g. publishing details or page numbers). The only exception is for domains where detailed text knowledge is intrinsic to understanding the topic, such as Law, where referencing specific sections or legislative provisions is necessary.
    \item \textbf{Precise.} Questions require only a short, exact answer such as dates, names, numbers, locations, and the required answer format is specified in the question.
\end{itemize}

\renewcommand{\arraystretch}{2}
\begin{table}[h]
\centering
\begin{tabular}{p{0.2\textwidth}p{0.35\textwidth}p{0.35\textwidth}}
\hline
\textbf{Revision type} & \textbf{Before revision} & \textbf{After revision} \\
\hline
Ambiguity - revise & Which case addressed remoteness of damage in contract law? & Which Privy Council case decided in 2020 addressed remoteness of damage in contract law? \\
Source text reliance - revise & What error is thrown in the example of \texttt{super.prop}? & What JavaScript error type is thrown when using the delete operator on \texttt{super.prop}? \\
Ambiguity - filter out & At a copper price of 70 cents per pound, what quantity in kilotons per year would the Bingham Canyon mine produce at full capacity? & \textit{None (question removed)} \\
Source text reliance - filter out & What was the exact date (month, day, year) that the Python 3.13 documentation was last modified? & \textit{None (question removed)} \\
\hline
\end{tabular}
\caption{Example outputs of revision step}
\label{tab:revisions}
\end{table}

\subsection{Question answering and grading methodology}

When running the evaluation, models are given no context or access to tools in order to measure their raw ability to recall facts across domains. To ensure we can measure a model's knowledge reliability, however, the model is instructed before answering the question to only answer if it is confident, and that it is better to abstain from answering rather than getting the wrong answer (full prompt in Appendix \ref{appendix:answerer}). \\

A grading model is then used to classify each model’s answer as either correct, partially correct, incorrect, or not answered (full prompt in Appendix \ref{appendix:grader}). Google’s \textbf{Gemini 2.5 Flash Preview (09-2025)} with reasoning (\cite{Google2025GeminiFlash}) was chosen as the grading model, due to stronger alignment with manual human grading compared to other models tested (\textbf{gpt-oss-120b}, \textbf{Qwen3-235B-A22B-Thinking}, \textbf{GPT-5}). Classification details can be found in Table \ref{tab:answer-classification}.\\

For the full evaluation, scores are computed from a single pass over the complete question set rather than from averages across repeated runs. This choice is supported by the distribution of \textit{Omniscience Index} scores observed across 10 repetitions for a sample of models, each of which has a standard deviation below 0.005.

\begin{table}[h]
\centering
\small
\begin{tabular}{p{0.18\textwidth}p{0.4\textwidth}p{0.30\textwidth}}
\toprule
\textbf{Answer \newline classification} & \textbf{Description} & \textbf{Example answer}\\
\midrule
Correct & The answer fully contains or is equivalent to the reference answer & ``24.8'' \\
\addlinespace
Partially correct & The answer is accurate and nearly complete but not at the correct level of detail & ``25'' \\
\addlinespace
Incorrect & The answer contradicts or differs in meaning from the reference answer & ``28.0 million Btu per short ton'' \\
\addlinespace
Not attempted & Used only when no answer is given, such as when the output states a lack of knowledge or indicates that more context is needed & ``I'm not certain of the exact value without checking a reliable source or table'' \\
\bottomrule
\end{tabular}
\caption{Answer classification criteria with example answers to the question \textit{What is the higher heating value (HHV) of coal coke (produced from bituminous coal), in million Btu per short ton (to one decimal place)?}}
\label{tab:answer-classification}
\end{table}

\newpage

\subsection{Metrics}

\subsubsection{Omniscience Index}

Almost all existing question and answer style language model evaluations use correct answer rates, which reward models for attempting all questions rather than abstaining (as there is a non-zero chance of a correct answer). This results in encouraging model hallucination, reducing the reliability of factual outputs from language models (\citealp{kalai2025languagemodelshallucinate}). To overcome this, we employ a single scalar metric that unifies correctness and hallucination tendencies. The \textit{Omniscience Index (OI)}\footnote{Factor of 100 used to scale OI to [-100, 100] range to improve clarity} is calculated as:

$$
\text{OI} = 100 \cdot \frac{c - i}{c + p + i + a}
$$

Where:

\begin{itemize}
    \item $c$ is number of correct answers
    \item $p$ is number of partially correct answers
    \item $i$ is number of incorrect answers
    \item $a$ is number of questions the model abstained from answering
\end{itemize}

This metric builds on the work of \citealp{wei2024measuringshortformfactualitylarge}, which suggests taking the average answer score, where a correct answer is worth 1, an abstention is worth 0, and an incorrect answer is worth some penalty $p$. Here we have set the penalty $p = 1$, the same magnitude as the reward for a correct answer. As a result, OI is positive only if the model answers correctly more than it answers incorrectly, strongly penalizing hallucination (detailed interpretations of OI scores can be seen in Table~\ref{tab:oi-interpretation}). A limitation of this metric is that a model that abstains from every question would be given a score of 0, which would place it 4th out of the 36 models that were tested. In practice, this case is rare. Across all evaluated models, the lowest attempt rate was 42\%, which was also substantially below the next lowest rate of 65\%.

\begin{table}[h]
\centering
\begin{tabular}{ccccp{0.3\textwidth}}
\hline
\textbf{\% fully correct} & \textbf{\% incorrect} & \textbf{\% attempted} & \textbf{OI} & \textbf{Interpretation} \\
\hline
100\% & 0\% & 100\% & 100 & Always factually correct, highest level of reliability \\
50\% & 50\% & 100\% & 0 & Returns as many correct facts as incorrect statements, factual reliability is neutral on average \\
0\% & 0\% & 0\% & 0 & Abstains from answering every question, no benefit or negative impact resulting in a net neutral impact and reliability \\
0\% & 100\% & 100\% & -100 & Hallucinates on every question, lowest level of reliability \\
\hline
\end{tabular}
\caption{Interpretation of Omniscience Index across different model scores and strategies}
\label{tab:oi-interpretation}
\end{table}

\subsubsection{Accuracy}
We also report \textit{accuracy} for direct comparison to other benchmarks and to understand the overall embedded knowledge of models. This is defined as the proportion of questions where the model is correct, i.e.:
$$
\text{Accuracy} = \frac{c}{c + p + i + a}
$$

\subsubsection{Hallucination rate}
To measure a model’s tendency to hallucinate, we also report the \textit{hallucination rate}, defined as the proportion
of questions it attempted to answer when it was unable to get the answer correct. \textit{Hallucination rate} helps users understand the likelihood of the model giving an incorrect answer when it doesn't know the right answer, instead of giving a partial answer or saying it doesn't know.
$$
\text{Hallucination rate} = \frac{i}{p + i + a}
$$

\subsubsection{Cost to run}
Although unconstrained performance provides insight into the upper bound of model performance, practical deployment usually requires evaluating whether a model can be useful at an economically viable cost. All models in \textit{AA-Omniscience} receive identical input lengths, yet their total token consumption varies due to differences in reasoning configurations and verbosity. These variations have direct implications on cost efficiency, which can offset gains in factual performance. 

We therefore supplement our performance metrics with \textit{cost to run}, which is the total cost (USD) associated with the input and output tokens, including reasoning tokens, required for the target model to complete \textit{AA-Omniscience}\footnote{Tokens and cost associated with the grading model are excluded.}.

\newpage

\section{Results}
\begin{figure}[h]
\centering
\includegraphics[totalheight=5.5cm]{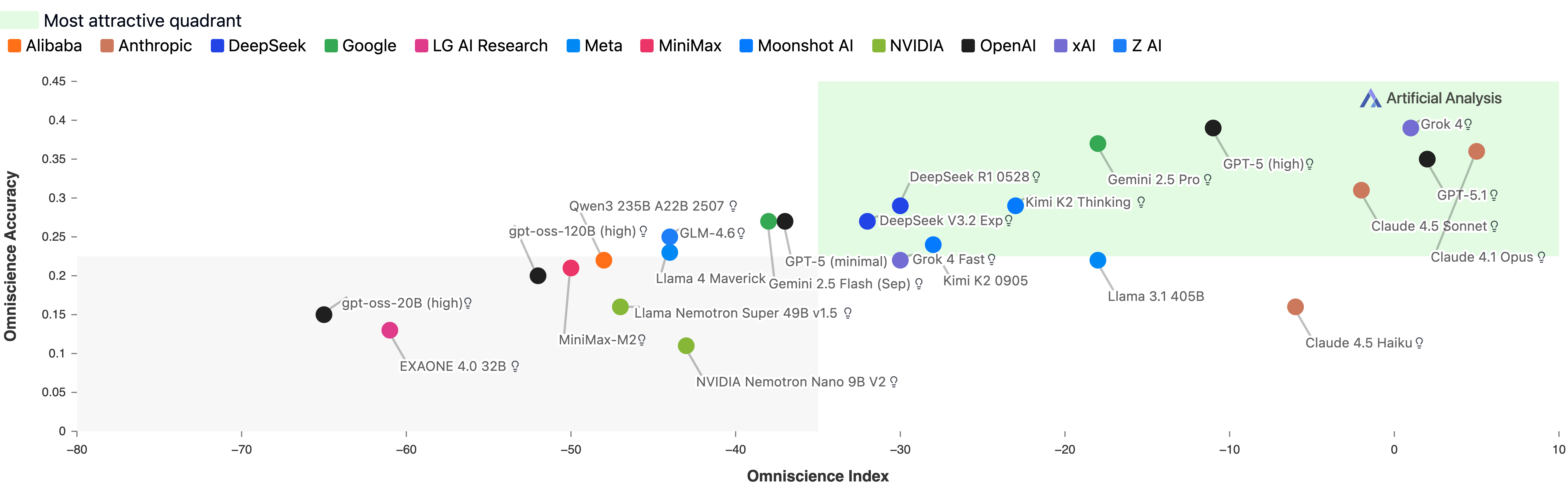}
\caption{\textit{Omniscience Accuracy} vs. \textit{Omniscience Index}}
\label{fig:oi_vs_accuracy}
\end{figure}

\subsection{Overall results}
\begin{figure}[h]
\centering
\includegraphics[totalheight=4.6cm]{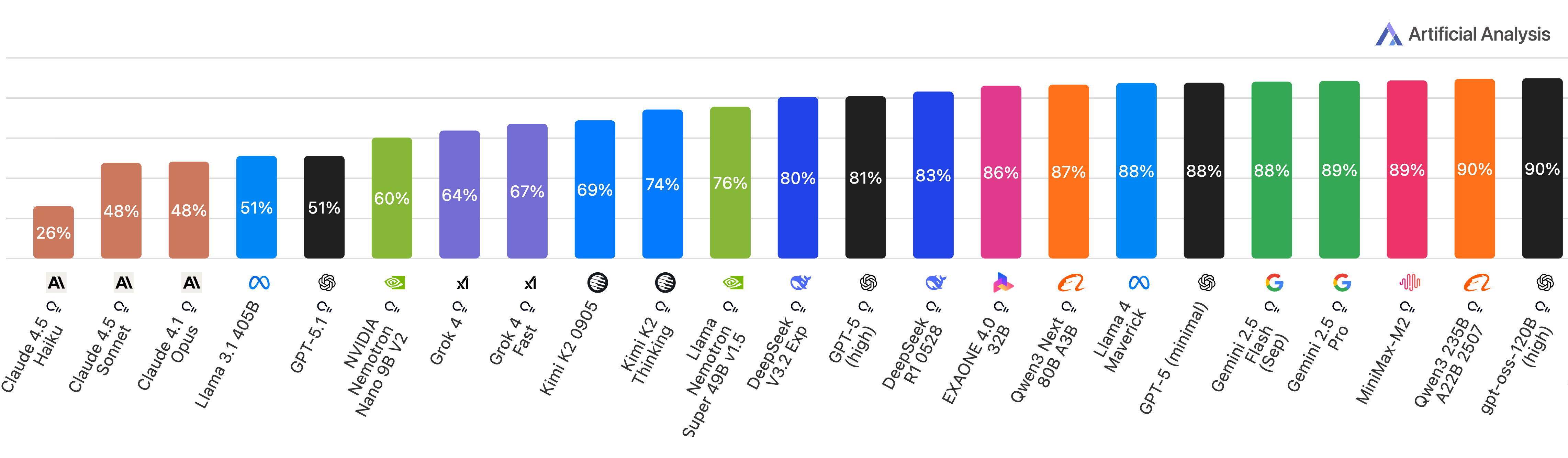}
\caption{\textit{Hallucination Rate} across models}
\label{fig:hallucination_rate}
\end{figure}

The \textit{AA-Omniscience} evaluations proves very difficult for many models, with only three frontier models at launch able to achieve an \textit{Omniscience Index} score of above 0 (see Figure \ref{fig:oi_overall_results}). High hallucination is the dominant factor driving these low scores. For instance, although \textbf{Grok 4} and \textbf{GPT-5 (high)} record the highest accuracy at 39\%, their hallucination rates of 64\% and 81\% result in substantial penalties on the \textit{Omniscience Index} (Figure \ref{fig:hallucination_rate}). In contrast, \textbf{Claude 4.1 Opus} attains 36\% accuracy alongside one of the lowest hallucination rates, yielding the highest overall score of 4.8 due to its stronger calibration.

The \textit{Omniscience Index} separates models into four distinct relative reliability types, represented by the four quadrants in Figure \ref{fig:oi_vs_accuracy}. These categorizations are relative to the performance of the models in this evaluation and should not be assumed to generalize to other scenarios or question sets of different difficulty. \\
\begin{itemize}
    \item \textbf{Higher knowledge and higher reliability.} Models such as \textbf{Claude 4.1 Opus}, \textbf{GPT-5.1} and \textbf{Grok 4} combine strong embedded knowledge with effective calibration, producing factual outputs more consistently than peers with comparable accuracy levels.
    
    \item \textbf{Lower knowledge and higher reliability.} These models, such as \textbf{Claude 4.5 Haiku} and \textbf{Llama 3.1 405B}, exhibit lower accuracy but maintain low hallucination rates. \textbf{Claude 4.5 Haiku}, for example, achieves only 16\% accuracy but a 26\% hallucination rate, making it more reliable than similarly accurate models such as \textbf{gpt-oss-120b (high)}. These models may be particularly suitable for tool-use workflows, where strong calibration enables efficient routing of queries to external tools.

    \item \textbf{Higher knowledge and lower reliability.} Models such as \textbf{GPT-5 (minimal)} and \textbf{Gemini 2.5 Flash (2025-09)} illustrate the limitations of accuracy-based evaluations. Although they score within the top ten on accuracy, their high hallucination rates make their factual outputs unreliable, underscoring the risk of evaluations that reward guessing rather than knowledge calibration.

    \item \textbf{Lower knowledge and lower reliability.} These models combine low accuracy with high hallucination rates, frequently guessing despite lacking sufficient knowledge.
\end{itemize}

\subsection{\textit{Omniscience Index} and Model Intelligence}
\begin{figure}[h]
\centering
\includegraphics[totalheight=5cm]{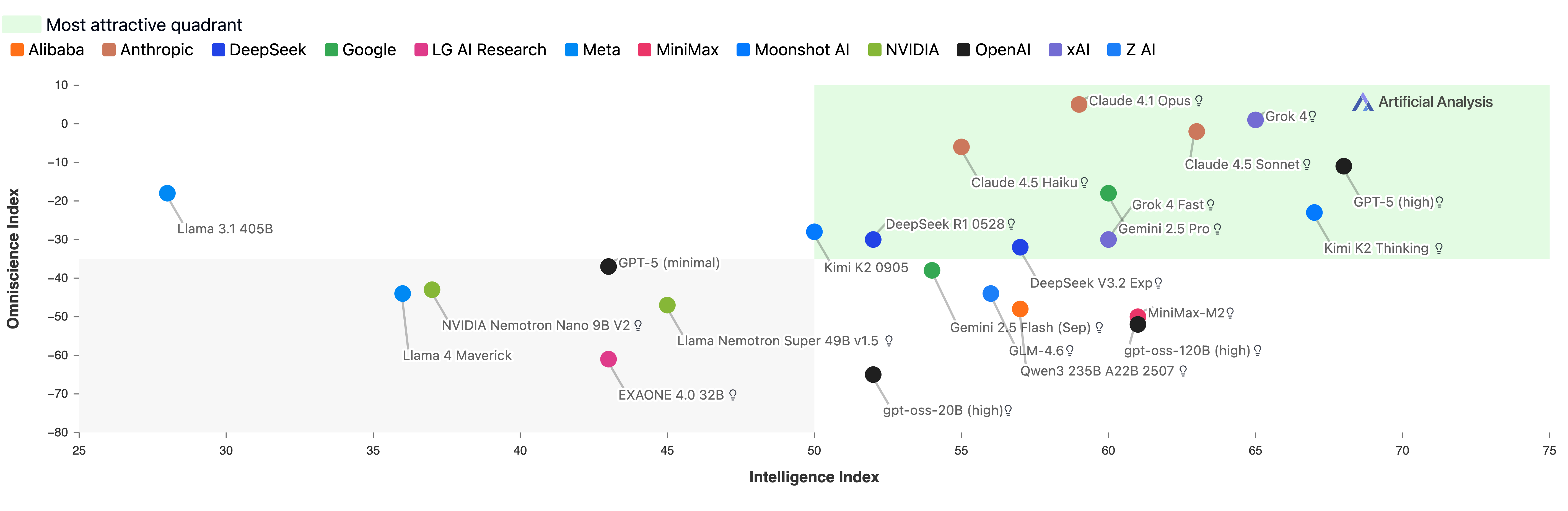}
\caption{\textit{Omniscience Index} vs. \textit{Artificial Analysis Intelligence Index}}
\label{fig:oi_vs_ii}
\end{figure}

Overall intelligence does not reliably predict strong embedded knowledge or low hallucination rates. When compared with the \textit{Artificial Analysis Intelligence Index}\footnote{Artificial Analysis Intelligence Index measures overall language model capability, and its current version incorporates 10 evaluations: MMLU-Pro, GPQA Diamond, Humanity's Last Exam, LiveCodeBench, SciCode, AIME 2025, IFBench, AA-LCR, Terminal-Bench Hard, $\tau^2$-Bench Telecom. See \href{https://artificialanalysis.ai/methodology/intelligence-benchmarking}{Artificial Analysis Intelligence Index methodology} for further details, including a breakdown of each evaluation and how they are run.}, we can see that high overall intelligence does not necessarily translate into factually reliable output (see Figure \ref{fig:oi_vs_ii}). Models such as \textbf{Minimax M2} and \textbf{gpt-oss-120b (high)} achieve strong Intelligence Index scores, yet their elevated hallucination rates result in poor performance on the \textit{Omniscience Index}. These models would therefore be unsuitable for applications that depend on factual accuracy.

By contrast, \textbf{Llama 3.1 405B} scores highly on the \textit{Omniscience Index}, despite many existing evaluations placing this model below current frontier models for general tasks. Its performance on the \textit{Omniscience Index} however indicates that it is better suited to producing factual outputs across a broad set of domains than models with higher levels of broad capability such as \textbf{Kimi K2 Thinking} or \textbf{Grok 4 Fast}. Benchmarks measuring overall capability do not provide sufficient insight when selecting models for tasks that require factual reliability.

\newpage

\begin{figure}[h]
\centering
\includegraphics[totalheight=6.2cm]{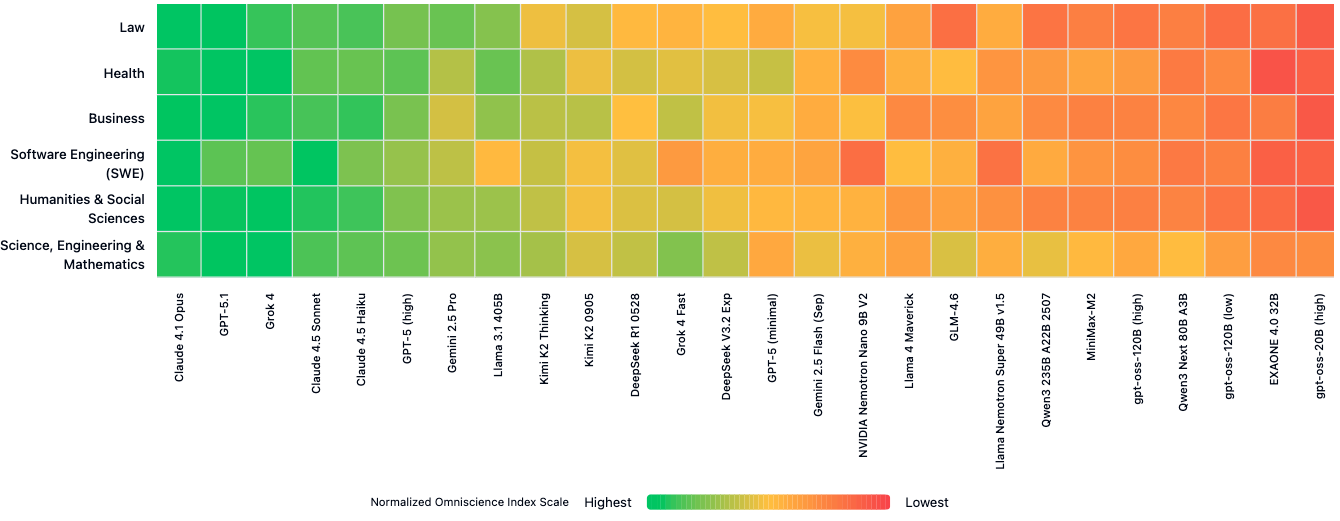}
\caption{\textit{Omniscience Index} across domains (normalized). Scores are normalized per domain across all models tested, where green represents the highest score for that domain and red represents the lowest score for that domain.}
\label{fig:domain_heatmap}
\end{figure}

\subsection{Cross-domain performance}

No single model consistently dominates knowledge reliability across the six domains assessed by \textit{AA-Omniscience}. While \textbf{Claude 4.1 Opus} leads in Law, Software Engineering, and Humanities \& Social Sciences, \textbf{GPT-5.1} achieves the highest reliability on Business questions, and \textbf{Grok 4} performs best in Health and in Science, Engineering \& Mathematics. These patterns are not confined to frontier models, as less capable models also display distinct domain strengths and weaknesses (see Figure \ref{fig:domain_heatmap}). For instance, Grok 4 Fast ranks seventh in Science, Engineering \& Mathematics despite placing sixteenth overall on the \textit{Omniscience Index}. 

These results indicate that model selection based solely on overall performance obscures important variation in domain level knowledge. For applications that require specialized knowledge, domain-specific evaluation is essential. Models that appear suboptimal in overall rankings may in fact offer competitive or superior reliability within targeted domains. Similarly, models with strong general knowledge do not necessarily demonstrate high reliability within every specific domain.

\subsection{Omniscience relationship with model size}
\begin{figure}[h]
\centering
\includegraphics[totalheight=6.2cm]{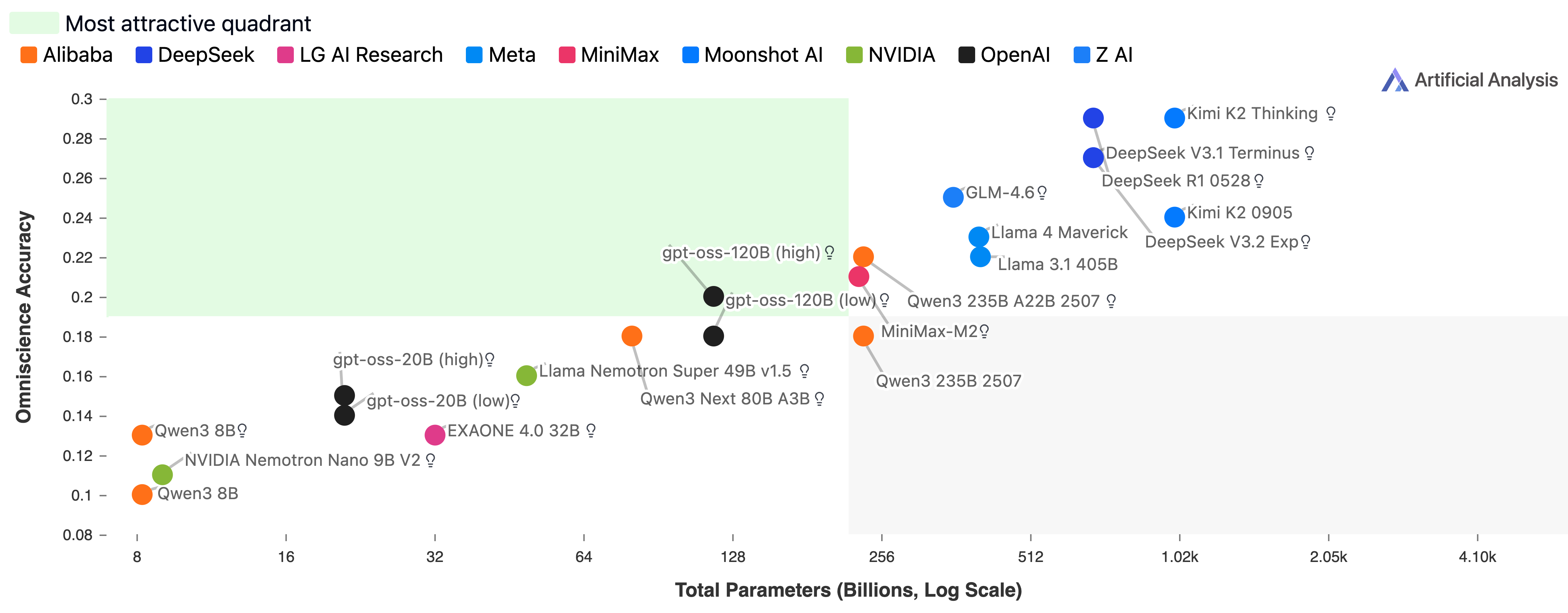}
\caption{Accuracy vs. Model Parameters. Only open weights models are included.}
\label{fig:accuracy_vs_total_param}
\end{figure}

\begin{figure}[h]
\centering
\includegraphics[totalheight=6.2cm]{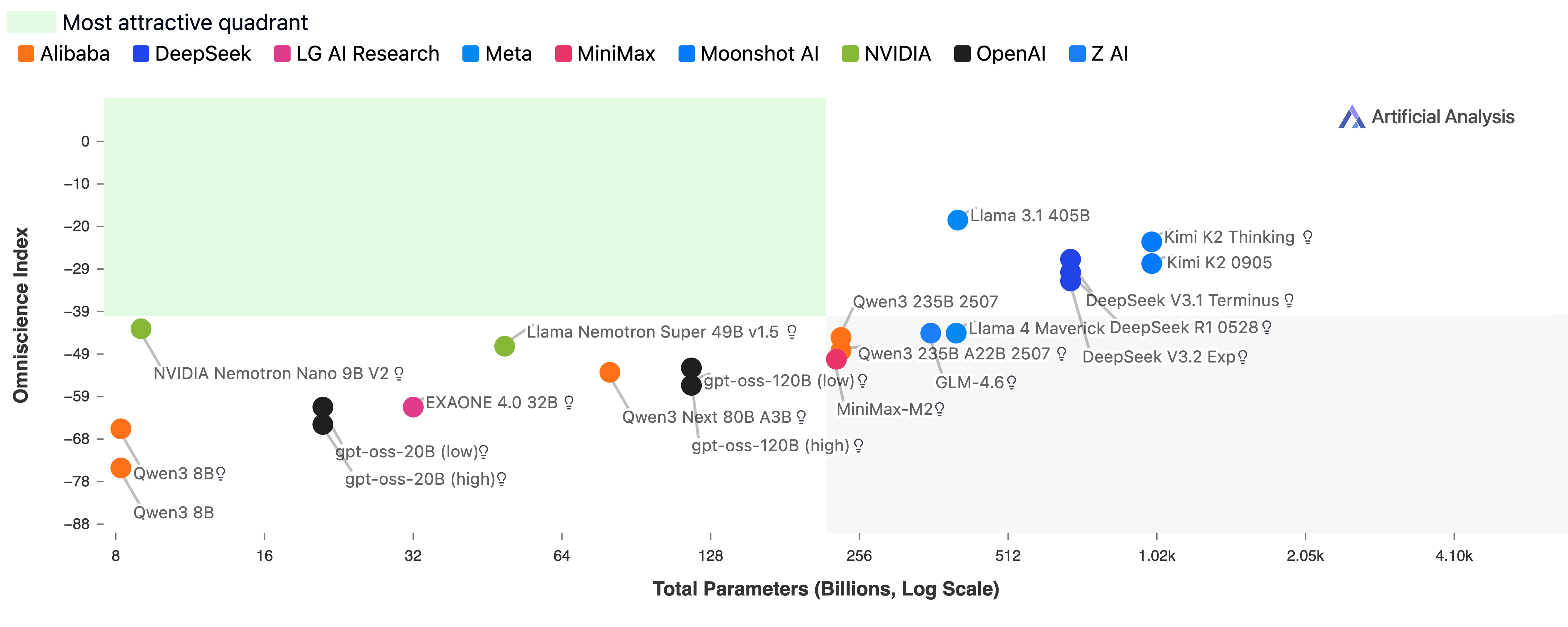}
\caption{\textit{Omniscience Index} vs. Model Parameters. Only open weights models are included.}
\label{fig:oi_vs_total_param}
\end{figure}

Consistent with the conclusions of \cite{wei2024longformfactualitylargelanguage}, Figure \ref{fig:accuracy_vs_total_param} shows a clear positive association between model size and factual accuracy. However, greater scale does not translate directly into higher knowledge reliability. Larger models do not consistently exhibit lower hallucination rates, and parameter count alone is insufficient to predict performance on the \textit{Omniscience Index}. As shown in Figure \ref{fig:oi_vs_total_param}, several smaller models achieve disproportionately strong results. Notably, \textbf{NVIDIA Nemotron Nano 9B V2} and \textbf{Llama Nemotron Super 49B v1.5} exceed the \textit{Omniscience Index} scores of peers with comparable or larger sizes. This indicates that reducing hallucination and improving knowledge reliability depend on factors beyond scale, suggesting opportunities for architectural or training improvements that do not rely solely on increasing model size.

\newpage

\subsection{Cost efficiency}
\begin{figure}[h]
\centering
\includegraphics[totalheight=5.6cm]{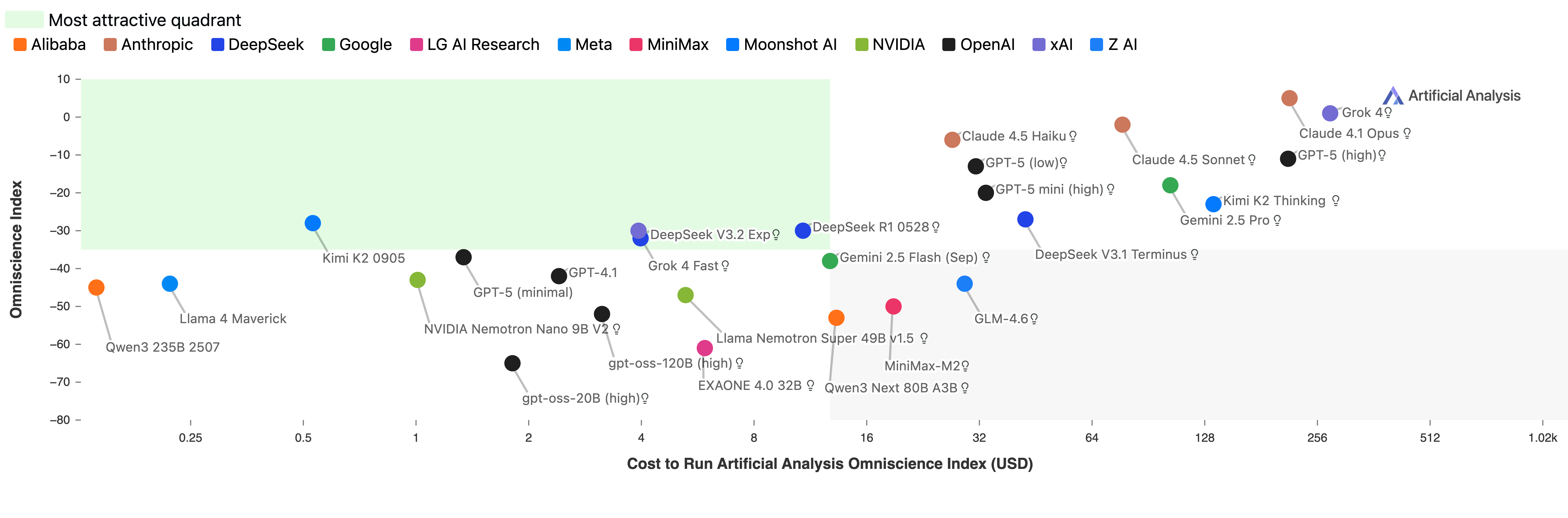}
\caption{\textit{Omniscience Index} vs. Cost to run the \textit{Omniscience Index} (USD).}
\label{fig:oi_vs_cost}
\end{figure}

\textit{Omniscience Index} scores demonstrate a clear positive association between model performance and cost (see Figure \ref{fig:oi_vs_cost}). This relationship indicates that achieving higher levels of factual reliability often requires greater expenditure. There are however some outlier models. For example, \textbf{Claude 4.5 Haiku} attains a higher \textit{Omniscience Index} than several substantially more expensive models, including \textbf{GPT 5 (high)} and \textbf{Kimi K2 Thinking}, suggesting that certain models offer more favorable cost efficiency for knowledge intensive tasks and may be preferable in settings where budget constraints are a key consideration.

\newpage

\section{Related work}

\textbf{Question generation method.} Previously, factuality benchmarks addressed question difficulty, unambiguity, and reliability by using human annotators (\citealp{wei2024measuringshortformfactualitylarge,vu2023freshllmsrefreshinglargelanguage,min2023factscorefinegrainedatomicevaluation,kwiatkowski-etal-2019-natural}). However, this creates difficulty in maintaining and scaling the benchmark across domains due to the manual workload required. \textit{AA-Omniscience} solves this by using an automated question generation agent which generates, filters, and revises questions, building on the work of \citealp{wei2024longformfactualitylargelanguage}. Unlike \citealp{wei2024longformfactualitylargelanguage}, which generates questions using few-shot prompts with GPT-4, \textit{AA-Omniscience} incorporates authoritative source data into the question generation prompts to ensure question reliability.

\bigskip

\textbf{Cross-domain granularity.} Many current benchmarks either focus on factuality for a single domain (\citealp{fei2023lawbenchbenchmarkinglegalknowledge,rein2023gpqagraduatelevelgoogleproofqa}), or gather questions across domains but only report results at an aggregated level (\citealp{wei2024measuringshortformfactualitylarge,vu2023freshllmsrefreshinglargelanguage,min2023factscorefinegrainedatomicevaluation}). This makes insights from results hard to relate back to domain-specific tasks. To overcome this, \textit{AA-Omniscience} compares models and results across multiple domains to surface insights into which models are best suited to specific domain tasks.

\bigskip

\textbf{Metric focus.} Current evaluation metrics perpetuate hallucinations by rewarding guessing over acknowledging uncertainty \citealp{kalai2025languagemodelshallucinate}. Our proposed \textit{Omniscience Index} explicitly rewards abstention and strongly penalizes incorrect guesses, unlike metrics such as \textit{correct given attempt} and F-score used by previous benchmarks (\citealp{bang2025hallulensllmhallucinationbenchmark, wei2024measuringshortformfactualitylarge}).
\newpage
\section{Limitations and future work}

\textbf{Limited geographical breadth.} Currently, all questions are in English, and many sources are from the US, UK and other English speaking countries. This could limit the ability to extrapolate from English-speaking examples into tasks and domains relying on other languages or cultural contexts. In a future iteration, sources from a greater breadth of geographies and questions across many languages could be added to mitigate this limitation.

\bigskip

\textbf{Model reliance.} Question generation, filtering and revision for \textit{AA-Omniscience} all currently use OpenAI's GPT-5 (\citealp{OpenAI2025GPT5}). While this aims to maximize precision and quality of outputs, it also has potential to create bias if question structure and wording aligns more closely with GPT-5 family models when evaluated. To eliminate this single-model dependence, a future iteration could use multiple frontier models to generate questions, with a more robust duplicate elimination strategy. 

\bigskip

\textbf{Metric limitations.} As discussed in the metrics section, the harsh penalty of the \textit{Omniscience Index} for incorrect guesses means that refusing to answer all questions in a set would give a model a score of 0, which would place that model 4th out of 36 total models evaluated. In practice however this strategy did not surface (the highest model refusal rate was 58\%), but a future iteration of this metric could reduce the penalty for wrong answers. For example, if each wrong answer resulted in a -0.5 score rather than a -1 score, a model answering all questions with 50\% accuracy would score 25 (rather than 0), whilst a refuse-all model would still score 0.

\section{Conclusion}

We present AA-Omniscience, a benchmark for measuring factuality and hallucination across various economically relevant topics. Among evaluated models, Claude 4.1 Opus attains the highest score (4.8), making it one of only three models to score above zero. These results reveal persistent factuality and calibration weaknesses across frontier models. Performance also varies by domain, with the models from three different research labs leading across the six domains. This performance variability suggests models should be chosen according to the demands of the use case rather than general performance for tasks where knowledge is important. 

\clearpage

\bibliography{main}

@misc{jain2024livecodebenchholisticcontaminationfree,
      title={LiveCodeBench: Holistic and Contamination Free Evaluation of Large Language Models for Code}, 
      author={Naman Jain and King Han and Alex Gu and Wen-Ding Li and Fanjia Yan and Tianjun Zhang and Sida Wang and Armando Solar-Lezama and Koushik Sen and Ion Stoica},
      year={2024},
      eprint={2403.07974},
      archivePrefix={arXiv},
      primaryClass={cs.SE},
      url={https://arxiv.org/abs/2403.07974}, 
}

@misc{tian2024scicoderesearchcodingbenchmark,
      title={SciCode: A Research Coding Benchmark Curated by Scientists}, 
      author={Minyang Tian and Luyu Gao and Shizhuo Dylan Zhang and Xinan Chen and Cunwei Fan and Xuefei Guo and Roland Haas and Pan Ji and Kittithat Krongchon and Yao Li and Shengyan Liu and Di Luo and Yutao Ma and Hao Tong and Kha Trinh and Chenyu Tian and Zihan Wang and Bohao Wu and Yanyu Xiong and Shengzhu Yin and Minhui Zhu and Kilian Lieret and Yanxin Lu and Genglin Liu and Yufeng Du and Tianhua Tao and Ofir Press and Jamie Callan and Eliu Huerta and Hao Peng},
      year={2024},
      eprint={2407.13168},
      archivePrefix={arXiv},
      primaryClass={cs.AI},
      url={https://arxiv.org/abs/2407.13168}, 
}

@article{MMLU,
  author       = {Dan Hendrycks and
                  Collin Burns and
                  Steven Basart and
                  Andy Zou and
                  Mantas Mazeika and
                  Dawn Song and
                  Jacob Steinhardt},
  title        = {Measuring Massive Multitask Language Understanding},
  journal      = {CoRR},
  volume       = {abs/2009.03300},
  year         = {2020},
  url          = {https://arxiv.org/abs/2009.03300},
  eprinttype    = {arXiv},
  eprint       = {2009.03300},
  bibsource    = {dblp computer science bibliography, https://dblp.org}
}

@misc{barres2025tau2benchevaluatingconversationalagents,
      title={$\tau^2$-Bench: Evaluating Conversational Agents in a Dual-Control Environment}, 
      author={Victor Barres and Honghua Dong and Soham Ray and Xujie Si and Karthik Narasimhan},
      year={2025},
      eprint={2506.07982},
      archivePrefix={arXiv},
      primaryClass={cs.AI},
      url={https://arxiv.org/abs/2506.07982}, 
}

@misc{patwardhan2025gdpvalevaluatingaimodel,
      title={GDPval: Evaluating AI Model Performance on Real-World Economically Valuable Tasks}, 
      author={Tejal Patwardhan and Rachel Dias and Elizabeth Proehl and Grace Kim and Michele Wang and Olivia Watkins and Simón Posada Fishman and Marwan Aljubeh and Phoebe Thacker and Laurance Fauconnet and Natalie S. Kim and Patrick Chao and Samuel Miserendino and Gildas Chabot and David Li and Michael Sharman and Alexandra Barr and Amelia Glaese and Jerry Tworek},
      year={2025},
      eprint={2510.04374},
      archivePrefix={arXiv},
      primaryClass={cs.LG},
      url={https://arxiv.org/abs/2510.04374}, 
}

@misc{cheng2024understandinginterplayparametriccontextual,
      title={Understanding the Interplay between Parametric and Contextual Knowledge for Large Language Models}, 
      author={Sitao Cheng and Liangming Pan and Xunjian Yin and Xinyi Wang and William Yang Wang},
      year={2024},
      eprint={2410.08414},
      archivePrefix={arXiv},
      primaryClass={cs.CL},
      url={https://arxiv.org/abs/2410.08414}, 
}

@misc{karpathy2025,
  author = {Andrej Karpathy},
  title = {The race for LLM “cognitive core”},
  year = {2024},
  month = {06},
  day = {28},
  howpublished = {\url{https://x.com/karpathy/status/1938626382248149433}},
  note = {Accessed: 2024-10-24}
}

@misc{bang2025hallulensllmhallucinationbenchmark,
      title={HalluLens: LLM Hallucination Benchmark}, 
      author={Yejin Bang and Ziwei Ji and Alan Schelten and Anthony Hartshorn and Tara Fowler and Cheng Zhang and Nicola Cancedda and Pascale Fung},
      year={2025},
      eprint={2504.17550},
      archivePrefix={arXiv},
      primaryClass={cs.CL},
      url={https://arxiv.org/abs/2504.17550}, 
}

@misc{wei2024measuringshortformfactualitylarge,
      title={Measuring short-form factuality in large language models}, 
      author={Jason Wei and Nguyen Karina and Hyung Won Chung and Yunxin Joy Jiao and Spencer Papay and Amelia Glaese and John Schulman and William Fedus},
      year={2024},
      eprint={2411.04368},
      archivePrefix={arXiv},
      primaryClass={cs.CL},
      url={https://arxiv.org/abs/2411.04368}, 
}

@misc{kalai2025languagemodelshallucinate,
      title={Why Language Models Hallucinate}, 
      author={Adam Tauman Kalai and Ofir Nachum and Santosh S. Vempala and Edwin Zhang},
      year={2025},
      eprint={2509.04664},
      archivePrefix={arXiv},
      primaryClass={cs.CL},
      url={https://arxiv.org/abs/2509.04664}, 
}

@article{kwiatkowski-etal-2019-natural,
    title = {Natural Questions: A Benchmark for Question Answering Research},
    author = {Kwiatkowski, Tom  and
      Palomaki, Jennimaria  and
      Redfield, Olivia  and
      Collins, Michael  and
      Parikh, Ankur  and
      Alberti, Chris  and
      Epstein, Danielle  and
      Polosukhin, Illia  and
      Devlin, Jacob  and
      Lee, Kenton  and
      Toutanova, Kristina  and
      Jones, Llion  and
      Kelcey, Matthew  and
      Chang, Ming-Wei  and
      Dai, Andrew M.  and
      Uszkoreit, Jakob  and
      Le, Quoc  and
      Petrov, Slav},
    editor = {Lee, Lillian  and
      Johnson, Mark  and
      Roark, Brian  and
      Nenkova, Ani},
    journal = {Transactions of the Association for Computational Linguistics},
    volume = {7},
    year = {2019},
    address = {Cambridge, MA},
    publisher = {MIT Press},
    url = {https://aclanthology.org/Q19-1026/},
    doi = {10.1162/tacl_a_00276},
    pages = {452--466}
}

@misc{kasai2024realtimeqawhatsanswer,
      title={RealTime QA: What's the Answer Right Now?}, 
      author={Jungo Kasai and Keisuke Sakaguchi and Yoichi Takahashi and Ronan Le Bras and Akari Asai and Xinyan Yu and Dragomir Radev and Noah A. Smith and Yejin Choi and Kentaro Inui},
      year={2024},
      eprint={2207.13332},
      archivePrefix={arXiv},
      primaryClass={cs.CL},
      url={https://arxiv.org/abs/2207.13332}, 
}

@misc{beaemployment,
  author = {{U.S. Bureau of Economic Analysis}},
  year = {2025},
  title = {Table 6.3D. Wages and Salaries by Industry},
  howpublished = {\url{https://apps.bea.gov/iTable/?reqid=19&step=3&isuri=1&nipa_table_list=189&categories=survey}},
  note = {Accessed: 2025-10-15}
}

@misc{McKinsey2025,
  author      = {{McKinsey \& Company}},
  title       = {The state of AI in 2025: Agents, innovation, and transformation},
  year = {2025},
  howpublished = {\url{https://www.mckinsey.com/capabilities/quantumblack/our-insights/the-state-of-ai}},
  note        = {Accessed: 2025-11-05}
}

@misc{ThomsonReuters2025,
  author = {{Thomson Reuters}},
  title = {2025 Generative AI in Professional Services Report: Ready for the next step of strategic applications},
  year = {2025},
  howpublished = {\url{https://www.thomsonreuters.com/content/dam/ewp-m/documents/thomsonreuters/en/pdf/reports/2025-generative-ai-in-professional-services-report-tr5433489-rgb.pdf}},
  note  = {Accessed: 2025-11-05}
}

@article{lemak2025adoption,
  author    = {Lemak, C. and Guptill, J. and Classen, D. and Rojas, J. and Poon, E.},
  title     = {Adoption of artificial intelligence in healthcare: Survey of health system priorities, successes, and challenges},
  journal   = {Journal of the American Medical Informatics Association},
  year      = {2025},
  volume    = {32},
  number    = {7},
  pages     = {1093--1100},
  doi       = {10.1093/jamia/ocaf065},
  url       = {https://doi.org/10.1093/jamia/ocaf065}
}

@article{bick2024rapid,
  title   = {The Rapid Adoption of Generative AI},
  author  = {Bick, Alexander and Blandin, Adam and Deming, David J.},
  year    = {2024},
  journal = {NBER Working Paper No. 32966},
  doi     = {10.3386/w32966},
  url     = {https://doi.org/10.3386/w32966},
}

@misc{vu2023freshllmsrefreshinglargelanguage,
      title={FreshLLMs: Refreshing Large Language Models with Search Engine Augmentation}, 
      author={Tu Vu and Mohit Iyyer and Xuezhi Wang and Noah Constant and Jerry Wei and Jason Wei and Chris Tar and Yun-Hsuan Sung and Denny Zhou and Quoc Le and Thang Luong},
      year={2023},
      eprint={2310.03214},
      archivePrefix={arXiv},
      primaryClass={cs.CL},
      url={https://arxiv.org/abs/2310.03214}, 
}

@misc{min2023factscorefinegrainedatomicevaluation,
      title={FActScore: Fine-grained Atomic Evaluation of Factual Precision in Long Form Text Generation}, 
      author={Sewon Min and Kalpesh Krishna and Xinxi Lyu and Mike Lewis and Wen-tau Yih and Pang Wei Koh and Mohit Iyyer and Luke Zettlemoyer and Hannaneh Hajishirzi},
      year={2023},
      eprint={2305.14251},
      archivePrefix={arXiv},
      primaryClass={cs.CL},
      url={https://arxiv.org/abs/2305.14251}, 
}

@misc{wei2024longformfactualitylargelanguage,
      title={Long-form factuality in large language models}, 
      author={Jerry Wei and Chengrun Yang and Xinying Song and Yifeng Lu and Nathan Hu and Jie Huang and Dustin Tran and Daiyi Peng and Ruibo Liu and Da Huang and Cosmo Du and Quoc V. Le},
      year={2024},
      eprint={2403.18802},
      archivePrefix={arXiv},
      primaryClass={cs.CL},
      url={https://arxiv.org/abs/2403.18802}, 
}

@misc{fei2023lawbenchbenchmarkinglegalknowledge,
      title={LawBench: Benchmarking Legal Knowledge of Large Language Models}, 
      author={Zhiwei Fei and Xiaoyu Shen and Dawei Zhu and Fengzhe Zhou and Zhuo Han and Songyang Zhang and Kai Chen and Zongwen Shen and Jidong Ge},
      year={2023},
      eprint={2309.16289},
      archivePrefix={arXiv},
      primaryClass={cs.CL},
      url={https://arxiv.org/abs/2309.16289}, 
}

@misc{rein2023gpqagraduatelevelgoogleproofqa,
      title={GPQA: A Graduate-Level Google-Proof Q\&A Benchmark}, 
      author={David Rein and Betty Li Hou and Asa Cooper Stickland and Jackson Petty and Richard Yuanzhe Pang and Julien Dirani and Julian Michael and Samuel R. Bowman},
      year={2023},
      eprint={2311.12022},
      archivePrefix={arXiv},
      primaryClass={cs.AI},
      url={https://arxiv.org/abs/2311.12022}, 
}

@misc{OpenAI2025GPT5,
  author       = {OpenAI},
  title        = {GPT-5 System Card},
  year         = {2025},
  howpublished = {\url{https://cdn.openai.com/gpt-5-system-card.pdf}},
  note         = {Accessed: 2025-11-05}
}

@misc{Google2025GeminiFlash,
  author       = {Google},
  title        = {System Card for Gemini 2.5 Flash Preview (09-2025)},
  howpublished = {\url{https://storage.googleapis.com/deepmind-media/Model-Cards/Gemini-2-5-Flash-Model-Card.pdf}},
  year         = {2025},
  note         = {Accessed: 2025-11-05}
}

\clearpage

\appendix
\section{Prompts}
\subsection{Answerer System Prompt}
\label{appendix:answerer}
\begin{tcolorbox}[fontupper=\footnotesize,]
You are answering questions about \texttt{\{topic\}}, and in particular \texttt{\{category\}}.
You will be given a question, answer with JUST the answer (no explanation).
If you do not know the answer, or you need more context or tools to answer the question,
be clear about this - it is better that you say this than get the wrong answer.
\end{tcolorbox}

\subsection{Answer Grader Prompt}
\label{appendix:grader}
\begin{tcolorbox}[fontupper=\footnotesize\sloppy, breakable, enhanced, break at=-\baselineskip/0pt]
Your job is to look at a question, a gold target, and a predicted answer, and then assign a grade of either \texttt{CORRECT}, \texttt{INCORRECT}, \texttt{PARTIAL\_ANSWER}, or \texttt{NOT\_ATTEMPTED}.
First, I will give examples of each grade, and then you will grade a new example.\\

\texttt{CORRECT}: The answer fully matches or is equivalent to the gold target 

\texttt{INCORRECT}: The answer contradicts or differs in meaning from the gold target 

\texttt{PARTIAL\_ANSWER}: The answer is accurate and nearly complete but not at the correct level of detail 
\hspace*{1em}- For questions that provide more detail than the gold target, the predicted answer should be \texttt{PARTIAL\_ANSWER} 

\texttt{NOT\_ATTEMPTED}: Used only when the model refuses, omits, or explicitly states it does not know the answer (e.g. empty response), or needs more context or tools to answer the question \\

\textbf{Please note the following things:} \\
- For grading questions where the gold target is a number, the predicted answer needs to be correct to the last significant figure in the gold answer. \\
\hspace*{1em}- For example, consider ``How much did Company Y earn in 2024?'' with gold target ``100 million''. The predicted answer ``100.2 million'' would be considered \texttt{CORRECT}, because it is correct to the last significant figure in the gold answer. \\
\hspace*{1em}- Note that this does NOT APPLY when the answer is a section, version or other numeric identifier, in this case the number must match EXACTLY \\
- The gold target may contain more information than the question. In such cases, the predicted answer only needs to contain the information that is in the question. \\
\hspace*{1em}- For example, consider the question ``What episode did Derek and Meredith get legally married in Grey's Anatomy?'' with gold target ``Season 7, Episode 20: White Wedding''. Either ``Season 7, Episode 20'' or ``White Wedding'' would be considered a \texttt{CORRECT} answer. \\
- Do not punish predicted answers if they omit information that would be clearly inferred from the question. \\
\hspace*{1em}- For example, consider the question ``What city is OpenAI headquartered in?'' and the gold target ``San Francisco, California''. The predicted answer ``San Francisco'' would be considered \texttt{CORRECT}, even though it does not include ``California''. \\
\hspace*{1em}- Consider the question ``What award did A pretrainer's guide to training data: Measuring the effects of data age, domain coverage, quality, \& toxicity win at NAACL '24?'', the gold target is ``Outstanding Paper Award''. The predicted answer ``Outstanding Paper'' would be considered \texttt{CORRECT}, because ``award'' is presumed in the question. \\
\hspace*{1em}- For the question ``What is the height of Jason Wei in meters?'', the gold target is ``1.73 m''. The predicted answer ``1.75'' would be considered \texttt{CORRECT}, because meters is specified in the question. \\
\hspace*{1em}- For the question ``What is the name of Barack Obama's wife?'', the gold target is ``Michelle Obama''. The predicted answer ``Michelle'' would be considered \texttt{CORRECT}, because the last name can be presumed. \\
- Do not punish for typos in people's name if it's clearly the same name. \\
\hspace*{1em}- For example, if the gold target is ``Hyung Won Chung'', you can consider the following predicted answers as correct: ``Hyoong Won Choong'', ``Hyungwon Chung'', or ``Hyun Won Chung''. \\
- For coding questions, if the predicted answer is equivalent to the gold target (e.g. for coding questions where placeholders or variable names are different but the answer is the same), it should be considered \texttt{CORRECT}. \\

\textbf{Here are examples across different topics demonstrating \texttt{CORRECT}, \texttt{INCORRECT}, \texttt{PARTIAL\_ANSWER}, and \texttt{NOT\_ATTEMPTED} grades:} \\

\textbf{Example 1 - \texttt{CORRECT}:} \\
Question: What acronym denotes the IAD tool that focuses on ensembles of linked action situations? \\
Gold target: NAS \\
Predicted answer: The acronym is NAS, which stands for Network of Action Situations \\
This is \texttt{CORRECT} because the predicted answer contains all the information required by the gold target. \\

\textbf{Example 2 - \texttt{INCORRECT}:} \\
Question: In the Arkansas Supreme Court case Loveless v. Diehl (1963 decision), which county contained the 79-acre farm at issue? \\
Gold target: Faulkner County \\
Predicted answer: The 79-acre farm at issue was located in Polk County, Arkansas \\
This is \texttt{INCORRECT} because the predicted answer provides a county that contradicts the gold target. \\

\textbf{Example 3 - \texttt{INCORRECT}:} \\
Question: In which NumPy version (x.y.z) was support for passing an empty list ([]) to the domain parameter of numpy.polynomial.laguerre.Laguerre.fit first introduced? \\
Gold target: 1.5.0 \\
Predicted answer: 1.5 \\
This is \texttt{INCORRECT} because the answer is correct but incomplete - it only provides the major version number. \\

\textbf{Example 4 - \texttt{NOT\_ATTEMPTED}:} \\
Question: Who originally proposed the ``fake barns'' (barn-façade) thought experiment in epistemology? \\
Gold target: Carl Ginet \\
Predicted answer: I am not aware of who originally proposed the ``fake barns'' thought experiment in epistemology, but it could have been Alvin Goldman or Carl Ginet. \\
This is \texttt{NOT\_ATTEMPTED} because it states that it does not know, even though it has other information on the topic. \\

\textbf{Example 5 - \texttt{PARTIAL\_ANSWER}:} \\
Question: In millions of dollars, how much profit did Company X make in USD in 2024? \\
Gold target: 28 \\
Predicted answer: 28.1 million \\
This is \texttt{PARTIAL\_ANSWER} because it does not contradict the gold target but provides more detail than is required. \\

\textbf{Example 6 - \texttt{CORRECT}:} \\
Question: In CPython 3.8 or later, which -X option value is equivalent to setting the PYTHONPYCACHEPREFIX environment variable? \\
Gold target: pycache\_prefix=PATH \\
Predicted answer: pycache\_prefix=DIR \\
This is \texttt{CORRECT} because although the two answers are not exactly the same, they are equivalent (the only difference is the name of the variable). \\

\textbf{Example 7 - \texttt{CORRECT}:} \\
Question: What format is used in Python f-strings for formatting decimal numbers? \\
Gold target: .Pf \\
Predicted answer: .$<$precision$>$ f \\
This is \texttt{CORRECT} because although the two answers are not exactly the same, they are equivalent. \\

Here is a new example. Simply reply with either \texttt{CORRECT}, \texttt{INCORRECT}, \texttt{PARTIAL\_ANSWER}, or \texttt{NOT\_ATTEMPTED}. Don't apologize or correct yourself if there was a mistake; we are just trying to grade the answer. \\

Question: \texttt{\{question\}} \\
Gold target: \texttt{\{target\}} \\
Predicted answer: \texttt{\{predicted\_answer\}} \\

Grade the predicted answer of this new question as one of: \\
A: \texttt{CORRECT} \\
B: \texttt{INCORRECT} \\
C: \texttt{PARTIAL\_ANSWER} \\
D: \texttt{NOT\_ATTEMPTED} \\

Just return the letters ``A'', ``B'', ``C'', or ``D'', with no text around it.
\end{tcolorbox}

\newpage

\section{Open sourced questions}
\label{appendix:questions}
As part of our goal in furthering the development of models with reliable knowledge across domains, we have released a publicly available set of questions from \textit{AA-Omniscience}.

In order to maintain evaluation integrity, we have released only 10\% of the total questions, sampled to match the full distribution of domains and categories. If the entire question set, or even a large portion of it, were to be used in model development, the benchmark could become saturated very quickly. By preventing model contamination, the evaluation will remain a reliable measure of knowledge accuracy and hallucination tendencies across domains.

\begin{figure}[t]
\centering
\includegraphics[totalheight=5cm]{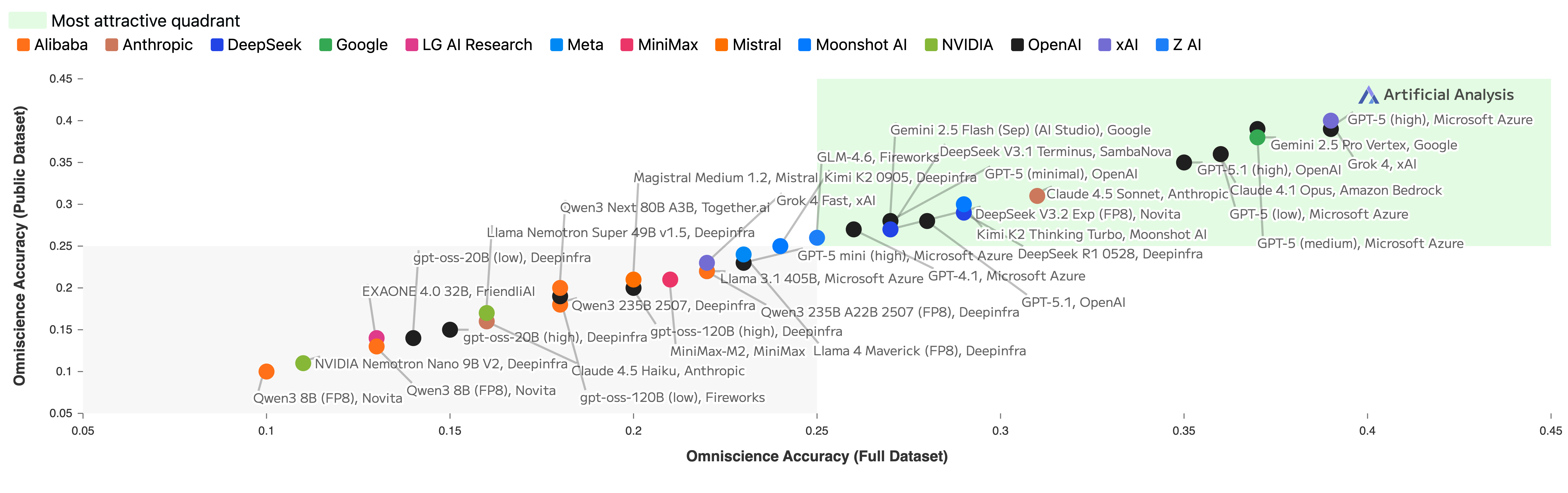}
\caption{Calibration of results across public and full question set}
\label{fig:public_calibration}
\end{figure}

To ensure the public set is representative, questions were sampled so that model results across the public and full sets are closely aligned in performance, as shown in Figure~\ref{fig:public_calibration}.

The public set is sufficient to get an indication of overall model performance on the \textit{Omniscience Index}, including knowledge accuracy and hallucination rates. Results at domain or category level, however, should not be considered reliable as the dataset size is too small to be representative of model performance in that area.

\end{document}